\documentclass[lettersize,journal]{IEEEtran}
\usepackage{amsmath,amsfonts}
\usepackage{algorithm}
\usepackage{algorithmicx}
\usepackage{algpseudocode}

\usepackage{array}
\usepackage[caption=false,font=normalsize,labelfont=sf,textfont=sf]{subfig}
\usepackage{textcomp}
\usepackage{stfloats}
\usepackage{multirow}
\usepackage{hyperref}
\usepackage{url}
\usepackage{verbatim}
\usepackage{graphicx}
\usepackage{cite}
\usepackage{booktabs}
\usepackage{xcolor} 

\begin{document}


\title{WaDiS: Boundary-Preserving Diffusion Segmentation via Multi-Scale Wavelet Conditioning and Frequency-Aware Gating}


\author{
Qiang Gao$^{1}$,
Yi Wang$^{2*}$,
Yong Zhang$^{2}$,
Bin Fang$^{2}$,
Yong Li$^{2}$,
Lan Du$^{1}$,
Cunjian Chen$^{1*}$\\[4pt]
{\normalsize
$^{1}$Department of Data Science and AI,
Monash University, Melbourne, Australia}\\
{\normalsize
$^{2}$College of Computer Science,
Chongqing University, Chongqing, China}\\[4pt]
{\normalsize
$^{*}$Corresponding authors:
yiwang@cqu.edu.cn;
Cunjian.Chen@monash.edu}
}



\maketitle 

\begin{abstract}
Diffusion models have shown strong potential in medical image segmentation due to their generative capabilities and robustness. However, existing methods often struggle to preserve fine anatomical boundaries and to retain multi-scale structural information due to the loss of spatial and frequency details during feature encoding, particularly in complex clinical imaging conditions. This work presents WaDiS, a wavelet-powered conditional latent diffusion framework for medical image segmentation that addresses these limitations through multi-scale wavelet conditioning and frequency-aware information filtering for boundary preservation. The model adopts a multi-scale wavelet conditioning encoder to retain rich structural features, leveraging its lossless nature to better preserve semantic priors critical to the diffusion process. In addition, a Frequency-Aware Gating Block (FAGB) is introduced into the skip connections of the diffusion U-Net, enabling adaptive spectral filtering and residual fusion to better emphasize boundary structures while reducing irrelevant content. To improve generalization across clinical domains, we adopt a structure-preserving color augmentation strategy during training. Specifically, global color statistics are transferred between samples in the LAB color space, allowing the model to decouple appearance from structure and focus more on anatomical shape rather than color bias. Experiments on five benchmark datasets demonstrate that WaDiS consistently outperforms existing diffusion-based and conventional segmentation methods, with notable improvements in boundary accuracy, robustness to noise, and generalization under color variation. The code for this work will be made publicly available upon acceptance.

\end{abstract}


\section{Introduction}
\begin{figure}[htbp]
\centerline{\includegraphics[scale=0.38]{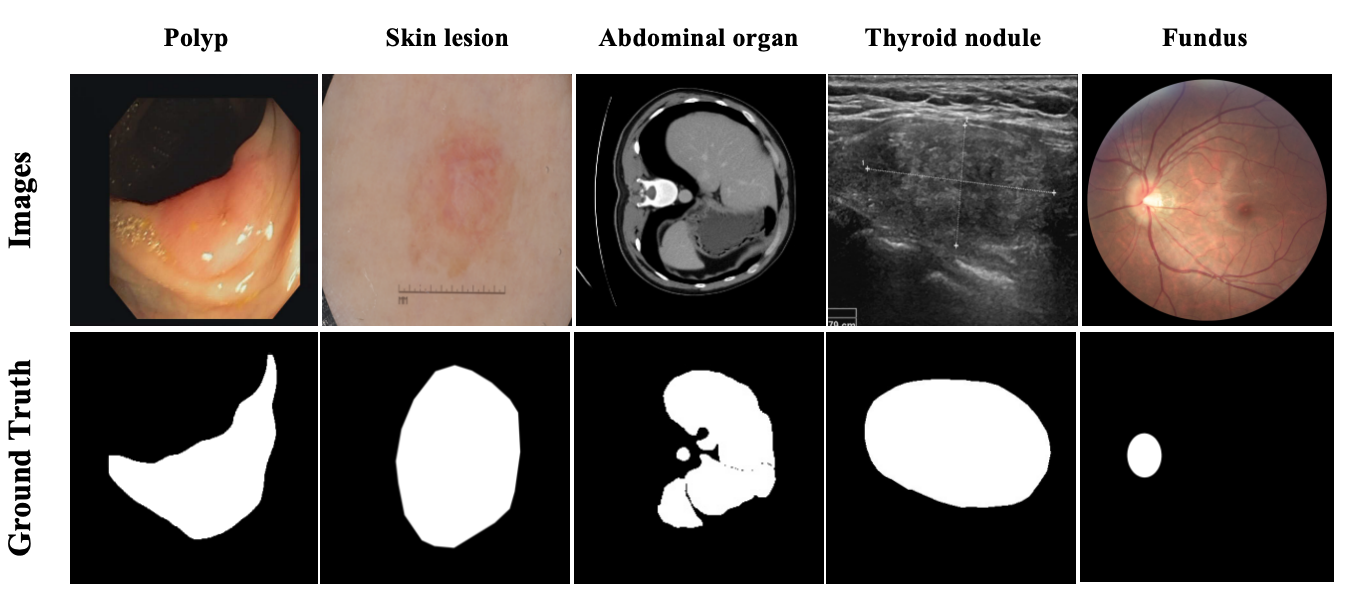}}
\caption{Representative examples from five medical imaging modalities, including (from left to right) colonoscopy, dermoscopy, CT, ultrasound, and fundus imaging. Even ground truth masks reflect uncertainty in boundary definition.}
\label{fig:boundary_challenge}
\end{figure}

\IEEEPARstart{A}{ccurate} segmentation of medical images~\cite{chen2024hidiff,rahman2024emcad,zhou2024bsbp, chen2023transattunet} is critical for the quantitative analysis of anatomical structures and pathological regions. It serves as a foundational step in numerous clinical workflows, including disease screening and diagnosis~\cite{azad2024advances, wang2022medical}, treatment planning~\cite{sherer2021metrics}, pre-operation planning~\cite{niu2021uterine}, surgical navigation~\cite{avrunin2015surgical}. Reliable segmentation directly influences clinical decision-making and plays a key role in advancing intelligent diagnosis systems and precision medicine through emerging computational intelligence techniques. Although expert annotation remains the standard for medical image segmentation, the process is inherently labor intensive and time consuming, requiring substantial domain knowledge and clinical expertise. These constraints pose significant challenges to scalability, particularly in high-throughput clinical environments or large-scale imaging studies. Furthermore, medical images often suffer from low contrast, heterogeneous textures, and fuzzy boundaries between anatomical structures and surrounding tissues~\cite{lei2025condseg,li2025kg}. Such characteristics make precise boundary delineation particularly challenging, even for modern deep learning-based segmentation models. These difficulties are further compounded in modalities such as ultrasound or CT, where noise, artifacts, and variable imaging conditions obscure critical semantic details~\cite{zhang2024low,gao2025useanet}. As shown in Fig.~\ref{fig:boundary_challenge}, anatomical structures frequently exhibit ambiguous boundaries and gradual intensity transitions, posing intrinsic challenges for both annotation and automated segmentation.

The aforementioned boundary ambiguities have motivated extensive research on discriminative segmentation frameworks, which learn direct mappings from medical images to pixel-level segmentation masks. Over the past decade, significant progress in medical image segmentation has been driven by supervised discriminative learning frameworks. Conventional methods are primarily based on convolutional neural networks (CNNs), such as U-Net and its numerous variants~\cite{unet,oktay2018attention,xiao2018weighted,zhou2018unet++,guan2019fully,isensee2021nnu}, which utilize encoder–decoder architectures with skip connections to capture hierarchical and multiscale features. While CNNs exhibit strong performance in modeling local spatial patterns, their inherently limited receptive fields and downsampling operations often lead to blurring of fine structural boundaries. To mitigate this limitation, recent studies have introduced hybrid architectures that incorporate self-attention mechanisms~\cite{vaswani2017attention} into convolutional backbones. Representative models such as TransUNet~\cite{transunet}, Transfuse~\cite{zhang2021transfuse}, and UTNet~\cite{gao2021utnet} combine the locality of convolutions with the global context modeling capabilities of transformers. While widely adopted, these models focus primarily on learning decision boundaries in the feature space without explicitly modeling the data distribution~\cite{chen2024hidiff}, which limits their ability to capture intrinsic class semantics. This often leads to unstable representations, making them vulnerable to performance degradation under boundary ambiguity, appearance shifts, or limited supervision.

Given the aforementioned limitations of discriminative segmentation models~\cite{rahman2024emcad,du2025net,lei2025condseg,zhu2024selfreg} in handling distribution shifts and capturing ambiguous boundaries, there is growing interest in generative approaches that model the underlying data distribution explicitly. Unlike discriminative models that solely focus on learning decision boundaries, generative frameworks aim to approximate the data manifold itself, thereby offering improved robustness under uncertainty. Among various generative paradigms, diffusion probabilistic models~\cite{ho2020denoising,nichol2021improved} have gained increasing attention due to their ability to model complex data distributions through iterative denoising processes. Initially introduced for high-quality image generation, these models have since been extended to tasks such as segmentation, owing to their ability to model complex data distributions and capture fine-grained data variability. By modeling image generation as a Markovian denoising process, diffusion models offer a principled framework for learning complex data distributions and have shown strong potential in maintaining spatial coherence in downstream tasks such as segmentation~\cite{wolleb2022diffusion,amit2021segdiff,wu2024medsegdiff}. In particular, SDSeg~\cite{lin2024stable} distinguishes itself by employing latent diffusion to significantly reduce computational overhead, while enabling efficient inference via single-step reverse generation and latent feature fusion. Despite these advantages, SDSeg still faces challenges in preserving fine anatomical structures and accurately delineating boundaries, particularly under low contrast or ambiguous visual conditions. To enhance the preservation of semantic and structural cues in the input image, we adopt a multi-scale wavelet decomposition, which offers a compact and informative representation well-suited for conditioning the diffusion model. Wavelet transforms~\cite{phung2023wavelet,korkmaz2024training} provide a compact and invertible multi-resolution representation, enabling the encoding of rich semantic information without any loss. In addition, leveraging frequency-domain filtering allows for the adaptive suppression of irrelevant spectral components while preserving critical boundary information. 

Motivated by the need for better structural awareness and boundary precision in latent diffusion models, we propose WaDiS, which combines a multi-scale wavelet condition encoder to capture hierarchical structure features with a boundary-aware gating mechanism that enhances boundary preservation throughout the generative process. Specifically, built upon a latent diffusion backbone, WaDiS introduces two core components: a wavelet-based encoder for multi-scale semantic conditioning, and a Frequency-Aware Gating Block (FAGB) embedded in the skip connections of the diffusion U-Net for structure-preserving feature refinement. Specifically, we replace the original condition encoder with a wavelet-driven feature extractor and embed FAGB modules into the skip connections of the diffusion U-Net. By leveraging lossless wavelet decomposition, the proposed condition encoder captures hierarchical structural cues to guide the generative process. Complementarily, the boundary-aware gating module improves boundary precision by combining frequency-domain filtering with residual gating. In addition to architectural improvements, we also incorporate a structure-preserving color augmentation strategy during training. This technique transfers global color statistics between real images in the LAB color space, encouraging the model to focus on anatomical structures rather than insignificant color information. It helps improve generalization across varied clinical environments where lighting conditions, imaging devices, and patient-specific factors may introduce significant color variation. We validate our approach on five benchmark datasets, where it consistently outperforms existing methods, especially in scenarios with ambiguous boundaries, noisy backgrounds, and varying appearance characteristics.

To summarize, our main contributions are as follows:

\begin{itemize}
    \item We propose a wavelet-powered latent diffusion architecture that leverages the invertible and multi-resolution nature of wavelet transforms for semantically rich and lossless conditioning.
    
    \item We introduce a Frequency-Aware Gating Block (FAGB) in the skip connections of the diffusion U-Net to enhance boundary preservation and suppress background noise through spectral filtering and residual fusion.

    \item We apply a structure-preserving color augmentation strategy during training, using LAB-based color statistics to improve generalization under varying clinical conditions.

    \item We validate our method on five medical segmentation benchmarks, achieving consistent improvements over existing diffusion-based and conventional approaches, especially in challenging scenarios.
\end{itemize}

\section{Related work}

\subsection{Diffusion-based Segmentation} 

In recent studies, diffusion models have shown significant promise across a wide range of segmentation tasks, demonstrating strong performance in the field of medical image segmentation~\cite{shuai2024diffseg,wu2024medsegdiff,chowdary2023diffusion,yan2024cold}. The diffusion process supports a conditional generative framework that progressively refines segmentation predictions by incorporating semantic and structural priors. Through a step-by-step denoising trajectory from random noise, diffusion-based segmentation models~\cite{chen2023berdiff,rahman2023ambiguous} often achieve better reconstruction quality. Recently, LEAF~\cite{huang2025leaf} framework enhances latent diffusion-based segmentation through feature distillation that aligns convolutional and transformer representations for better feature consistency and efficiency. In addition, cDAL~\cite{hejrati2024conditional} framework integrates spatial attention and latent embeddings into a conditional diffusion model, using a CNN-based discriminator to enhance segmentation of discriminative regions efficiently. Although these models have achieved encouraging results, the inherently complex anatomical structures and the low intensity contrast between adjacent tissues in medical images continue to challenge the accurate delineation of boundaries. Furthermore, many existing diffusion-based segmentation frameworks still rely on conventional CNN encoders or generic conditioning schemes~\cite{huang2025leaf}, which may not fully capture the structural diversity or the rich multi-scale frequency characteristics present in medical images.

\subsection{Wavelet-Based Feature Representation} 

Wavelet transforms have been extensively utilized in various image analysis tasks due to their ability to jointly encode spatial and frequency characteristics across multiple scales~\cite{jiang2023low, chen2024low, shi2025frequency}. This dual-domain representation is particularly beneficial for vision problems where both broad context and fine structural details must be preserved, such as in image enhancement~\cite{jiang2023low}, super-resolution~\cite{chen2024low}, and medical image segmentation~\cite{shi2025frequency, zhao2023wranet}. Unlike conventional convolutional encoders that tend to lose information through repeated downsampling, discrete wavelet transforms (DWT) allow for invertible hierarchical decomposition, enabling structure-preserving feature extraction and accurate information reconstruction. Recent studies have shown that wavelet-based architectures can improve performance in scenarios characterized by weak boundaries or low contrast, which are common in both natural and medical images~\cite{yadav2025samwave,zhao2023wranet, shi2025frequency}. Motivated by these properties, we replace the CNN-based encoder in SDSeg~\cite{lin2024stable} with a wavelet-based module in our WaDiS framework. This substitution eliminates information loss caused by downsampling and allows structurally rich multi-scale features to guide the diffusion process more effectively.

\subsection{Frequency-Aware Feature Enhancement for Boundary Preservation}
Frequency domain techniques~\cite{chang2024net} have long been used in image processing for denoising and feature enhancement due to their ability to separate structural information across spectral components. By decomposing images into frequency components, these methods allow selective manipulation of information, such as attenuating high-frequency noise or enhancing edge structures. In medical imaging, frequency analysis has been employed to improve boundary clarity and reduce irrelevant background patterns, particularly in scenarios involving low contrast or heterogeneous textures. Recent medical image segmentation methods~\cite{ruan2022mew,huang2021medical} have begun to explore frequency-aware modules to complement spatial-domain operations. Building upon this direction, we design a Frequency-Aware Gating Block (FAGB) that operates in the frequency domain. FAGB applies learnable complex filtering followed by gated residual fusion to suppress noise and reinforce anatomical boundaries. Integrated into the skip connections of our diffusion U-Net, this module facilitates anatomically consistent and boundary-preserving feature reconstruction.

\begin{table}[htbp]
\centering
\caption{Essential Notations with Corresponding Descriptions}
\label{tab:notations}
\begin{tabular}{ll}
\toprule
\textbf{Notation} & \textbf{Description} \\
\midrule
$I$ & Input medical image \\
$Y$ & Segmentation ground truth \\
$x_c = \phi(I)$ & Wavelet-based conditional embedding \\
$x_0 = \psi(Y)$ & Latent embedding encoded from ground truth \\
$x_T$ & Noised latent after $T$ Gaussian diffusion steps from $x_0$ \\
$\hat{Y} = \delta(\hat{x}_0)$ & Reconstructed output from latent prediction \\
$\text{DWT}(\cdot)$ & Discrete Wavelet Transform \\
$\text{IDWT}(\cdot)$ & Inverse Discrete Wavelet Transform \\
$\text{HREB}(\cdot)$ & High-Frequency Residual Enhancement Block \\
$\phi(\cdot)$ & Wavelet-based conditional encoder \\
$\psi, \delta$ & Fixed autoencoder pair for encoding/decoding \\
$\mathcal{F}(\cdot), \mathcal{F}^{-1}(\cdot)$ & 2D FFT and inverse FFT \\
$\text{Filter}(\cdot)$ & Learnable frequency-domain filter \\
$\text{FAGB}(\cdot)$ & Frequency-Aware Gating Block \\
$\text{DWC}(\cdot)$ & Depthwise Separable Convolution \\
$\mathcal{L}_{noise}$ & Noise prediction loss \\
$\mathcal{L}_{latent}$ & Latent reconstruction loss \\
$\mathcal{L}_{total}$ & Total training objective function\\
\bottomrule
\end{tabular}
\end{table}

\begin{figure*}[htbp]
\centering
\includegraphics[width=\textwidth]{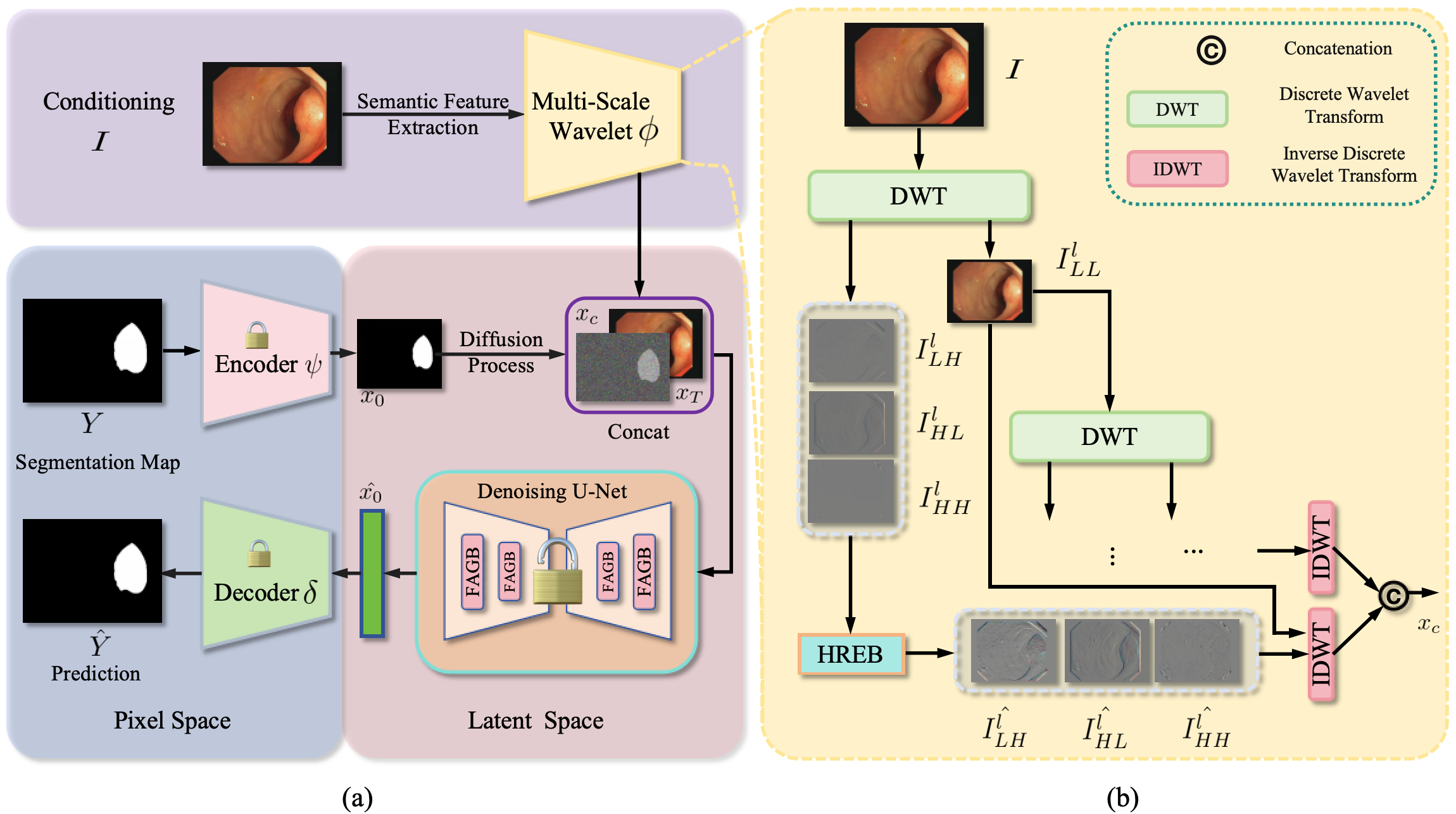}
\caption{Overview of the proposed WaDiS framework: (a) The overall architecture is built upon a latent diffusion model, where the segmentation process is formulated as a progressive denoising of the noised segmentation map’s latent representation. A multi-scale wavelet-based condition encoder extracts lossless embeddings from the input image to guide the denoising process, while Frequency-Aware Gating Blocks (FAGB) embedded in the skip connections of denoising U-Net enhance boundary-sensitive feature refinement; (b) The wavelet-based encoder decomposes the input image into hierarchical frequency sub-bands, providing multi-scale structural cues that serve as conditional guidance during latent denoising. High-Frequency Residual Enhancement Block (HREB) aims to strengthen the model’s sensitivity to texture and boundary information in the latent space.}
\label{fig2}
\end{figure*}

\subsection{Color Variation and Transfer Strategies for Robust Segmentation}
Color variation remains a persistent challenge in medical image segmentation, particularly due to differences in imaging protocols, devices, and post-processing procedures. To address this, color augmentation techniques~\cite{yan2025hvi,yan2024you} have become an essential component of training pipelines, enhancing model robustness and reducing overfitting by exposing the network to diverse color distributions. Data augmentation methods like ColorJitter are commonly used to address color variability and reduce overfitting problems~\cite{laskin2020reinforcement}. Earlier efforts in color transfer, such as the method by Reinhard et al.~\cite{reinhard2002color}, adapt image colors by matching global statistical characteristics between source and target images. The method proposed by HaCohen et al.~\cite{hacohen2011non}
ensures semantic consistency in color transfer by first establishing patch-level correspondences and then applying a global color mapping. Inspired by this idea, our color transfer operation aims to decouple color from semantic structure by altering global color tone, encouraging the model to focus on anatomical structure rather than dataset-specific appearance. 

\section{Methodology}

\subsection{Overall Architecture}
For medical image segmentation, we propose WaDiS, a latent diffusion framework enhanced with a wavelet-based conditional encoder and frequency-aware gating blocks embedded in the skip connections of the diffusion U-Net~\cite{unet}, as illustrated in Fig.~\ref{fig2}. In addition, a structure-preserving color augmentation strategy is applied during training to improve generalization across clinical domains, though it is not depicted in the figure.    

Given an input image $I \in \mathbb{R}^{H \times W \times 3}$, a latent representation $x_c = \phi(I)$ is first obtained using a condition encoder $\phi$, which replaces traditional encoders with a DWT-based approach that extracts multi-scale frequency features while retaining full structural fidelity through its lossless nature. The segmentation target $Y \in \mathbb{R}^{H \times W \times 3}$ is encoded into a latent mask $x_0 = \psi(Y)$ using a fixed autoencoder $(\psi, \delta)$ pretrained from a diffusion backbone, with $\hat{Y} = \delta(\hat{x_0})$ used for reconstruction. To ensure representational consistency and optimization stability, the pretrained autoencoder remains frozen throughout training. During training, the latent representation $x_0$ is progressively corrupted by adding Gaussian noise for $T$ time steps, resulting in a noisy latent $x_T$. The diffusion model is then trained to recover $x_0$ from $x_T$ conditioned on the latent input $x_c$, effectively learning a reverse denoising process that restores noisy latent features back to clean segmentation maps. The denoising process takes place entirely in the latent space via a U-Net network. To improve edge precision and suppress irrelevant noise, Frequency-Aware Gating Blocks (FAGB) are placed within the skip connections in the denoising U-Net. These modules apply learnable complex-valued filters followed by residual gating, selectively enhancing informative regions while filtering out undesired frequency components. To enhance robustness to color variability, we apply a structure-preserving color augmentation during training by transferring global LAB color statistics from a random image. By matching channel-wise means and standard deviations, the model learns to focus on anatomical structure over appearance. This transformation is applied only during training; inference uses original inputs. Additionally, Table \ref{tab:notations} provides the necessary
notations and the corresponding definitions for reference.

\subsection{Multi-Scale Wavelet Encoder for Conditional Embedding} 
In diffusion-based segmentation frameworks, the conditional encoder plays a crucial role in providing semantic priors to guide the generative process. This is particularly important in our setting, where the segmentation label is a binary mask with limited structural information. To enhance the encoder’s ability to retain structural fidelity~\cite{jiang2023low} and multi-scale context, we propose to integrate a wavelet-based multi-scale encoder as the conditional pathway, as illustrated in Fig.~\ref{fig2}(b) 

\begin{figure}[htbp]
\centerline{\includegraphics[scale=0.31]{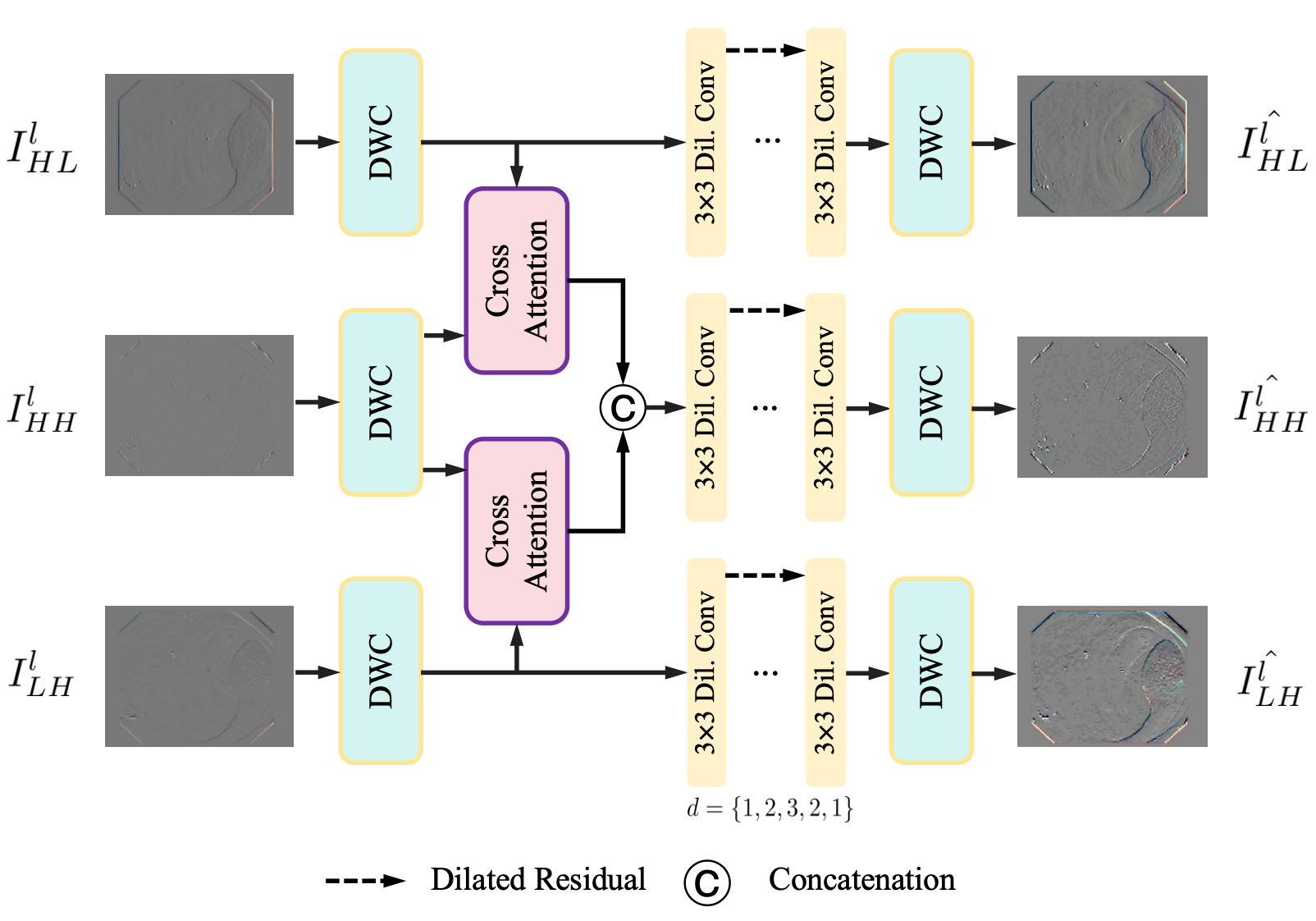}}
\caption{Design of the Proposed High-Frequency Residual Enhancement Block (HREB).}
\label{fig3}
\end{figure}

Given an input image $I \in \mathbb{R}^{H \times W \times 3}$, we perform a multi-level Discrete Wavelet Transform (DWT) to decompose the image into hierarchical frequency sub-bands. At each level $l$, the transformation yields four components: 
\begin{equation}
\{I_{LL}^l, I_{LH}^l, I_{HL}^l, I_{HH}^l\} = \text{DWT}(I),
\end{equation}

\begin{equation}
\{\hat{I_{HL}^l},\hat{I_{HH}^l},\hat{I_{LH}^l}\} = \text{HREB}(I_{HL}^l, I_{HH}^l,I_{LH}^l),
\end{equation}
where $I_{LL}^l$ is the approximation (low-frequency) coefficient at scale $l$, and the remaining terms represent horizontal, vertical, and diagonal high-frequency details. $\text{HREB}(\cdot)$ denotes the High-Frequency Residual Enhancement Block, which refines the high-frequency sub-bands to enhance edge and texture details.

The decomposition is recursively applied to $I_{LL}^l$ to extract deeper multi-scale representations:
\begin{equation}
\{I_{LL}^{l+1}, I_{LH}^{l+1}, I_{HL}^{l+1}, I_{HH}^{l+1}\} = \text{DWT}(I_{LL}^l),
\end{equation}

\begin{equation}
\{ \hat{I_{HL}^{l+1}}, \hat{I_{HH}^{l+1}}, \hat{I_{LH}^{l+1}}\} = \text{HREB}(I_{HL}^{l+1}, I_{HH}^{l+1},I_{LH}^{l+1}).
\end{equation}

We aggregate the wavelet sub-bands from two decomposition levels, forming the final condition embedding $\phi(I)$. The process is summarized as follows:
\begin{equation}
I' = \text{IDWT}(I_{LL}^{l},\ \hat{I_{LH}^l}, \hat{I_{HL}^l}, \hat{I_{HH}^l}),
\end{equation}

\begin{equation}
I'' = \text{IDWT}\left(I_{LL}^{l+1},\ \hat{I}_{LH}^{l+1},\ \hat{I}_{HL}^{l+1},\ \hat{I}_{HH}^{l+1}\right),
\end{equation}

\begin{equation}
\tilde{I}' = \text{Down}_{8\times}(I'), \quad
\tilde{I}'' = \text{Down}_{4\times}(I''),
\end{equation}

\begin{equation}
I''' = \text{Concat}(\tilde{I}',\ \tilde{I}''),
\end{equation}

\begin{equation}
\phi(I) = I''' = x_c,    
\end{equation}
where $\text{IDWT}(\cdot)$ is the inverse discrete wavelet transform, $\text{Down}{8\times}(\cdot)$ and $\text{Down}{4\times}(\cdot)$ denote downsampling by factors of 8 and 4, respectively, and $\text{Concat}(\cdot)$ indicates channel-wise concatenation. $x_c$ is the final wavelet-based conditional embedding.

The wavelet encoder brings two key benefits: it preserves the complete information of the input image through a lossless transformation, and it captures both coarse structures and fine details via multi-scale decomposition. Replacing the CNN backbone in the conditional branch with this encoder allows the diffusion process to operate under structurally-aware and multi-scale guidance.

\subsubsection{High-Frequency Residual Enhancement Block}

To better enhance structural boundaries and texture details, we introduce the High-Frequency Residual Enhancement Block (HREB), as illustrated in Fig.~\ref{fig3}. This block operates in the wavelet domain to refine high-frequency components at each decomposition level.

The high-frequency sub-bands contain localized edge and texture cues critical for accurate anatomical delineation. To enhance these features, we first apply depthwise separable convolutions to each high-frequency sub-band ${I_{LH}^l, I_{HL}^l, I_{HH}^l}$, followed by cross-attention blocks that aggregate complementary information from $I_{LH}^l$ and $I_{HL}^l$ into $I_{HH}^l$ to allow directional interaction and reinforce missing or weak responses in $I_{HH}^l$.

Next, we introduce a progressive dilated residual block, inspired by~\cite{hai2022combining}, consisting of sequential dilated convolutions with increasing and then decreasing dilation rates. This design balances local detail recovery with global context modeling, while mitigating the risk of gridding effect. The residual connection ensures stability and preserves the base representation. Finally, another set of depthwise separable convolutions is applied to compress the feature channels and produce the refined outputs ${\hat{I_{LH}^l}, \hat{I_{HL}^l}, \hat{I_{HH}^l}}$. 

\begin{figure}[htbp]
\centerline{\includegraphics[scale=0.24]{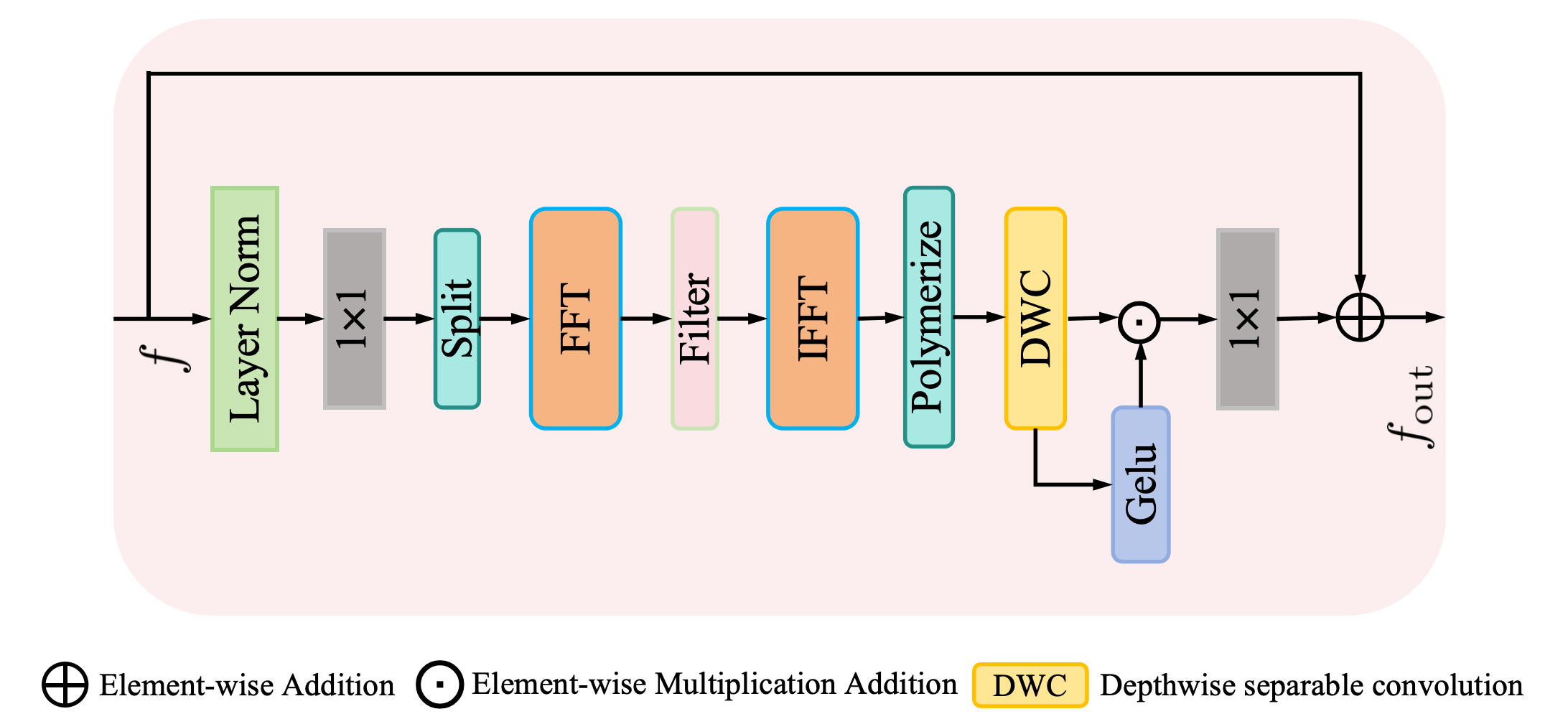}} 
\caption{Illustration of the Frequency-Aware Gating Block (FAGB).}
\label{fig4}
\end{figure}

\subsection{Frequency-Aware Gating Block (FAGB)}

In medical image segmentation, challenges arise from the need to distinguish anatomical boundaries in the presence of low contrast and visual noise. To address this, we design a Frequency-Aware Gating Block (FAGB), illustrated in Fig.~\ref{fig4}, which refines latent features by adaptively enhancing semantically important components in the frequency domain. This block is placed in the skip connections of the denoising U-Net, where it guides boundary-aware feature refinement during the reverse diffusion process.

Given an input feature map $f \in \mathbb{R}^{H \times W \times C}$, we first apply layer normalization followed by a $1 \times 1$ convolution to expand the channel dimension. This promotes richer representations and prepares the features for frequency projection. The output is then partitioned and transformed using a 2D Fast Fourier Transform~\cite{he2025frequency} (FFT), yielding a complex-valued representation $\hat{f}$ that separates spatial information across frequency components. To selectively emphasize meaningful structures, we apply a learnable complex filter consisting of two parallel convolution layers: one for the real part and one for the imaginary part. The filters are initialized close to an identity mapping to preserve frequency integrity while allowing fine-tuned adaptation during training. This filtering operation suppresses noise and redundant components, while enhancing texturally and structurally salient information across frequency and spatial axes.

After filtering, the processed features are transformed back to the spatial domain via inverse FFT. A depthwise separable convolution (denoted DWC) is then used to enhance local representations at the pixel level. Next, the feature map is divided into two channels. One branch is passed through a GELU activation, and its output is then multiplied element-wise with the other branch to produce a selectively gated response. A final $1 \times 1$ convolution reduces the channel dimension, and a residual connection adds the refined features back to the original input. Algorithm~\ref{alg:fagb} summarizes the computation steps of the Frequency-Aware Gating Block (FAGB).

\begin{algorithm}
\caption{Frequency-Aware Gating Block (FAGB)}
\label{alg:fagb}
\begin{algorithmic}[1]
\State $f_1 \gets \mathrm{LayerNorm}(f)$
\State $f_2 \gets \text{Conv}_{1\times1}(f_1)$
\State $f_3 \gets \text{Split}(f_2)$
\State $\hat{f} \gets \mathcal{F}(f_3)$ 
\State $\hat{f}_{\text{filtered}} \gets \text{Filter}(\hat{f})$ 
\State $f_4 \gets \mathcal{F}^{-1}(\hat{f}_{\text{filtered}})$ 
\State $f_5 \gets \text{Polymerize}(f_4)$ 
\State $f_6 \gets \text{DWC}(f_5)$ 
\State $f_7 \gets \mathrm{Gelu}(f_6) \odot f_6$ 
\State $f_8 \gets \text{Conv}_{1\times1}(f_7)$
\State \Return $f_{\text{out}} \gets f + f_8$ 
\end{algorithmic}
\end{algorithm}

\noindent
Notation in Algorithm~\ref{alg:fagb}:
$\mathrm{LayerNorm}(\cdot)$ applies per-channel normalization.
$\text{Conv}_{1\times1}(\cdot)$ denotes a $1\times1$ convolution used for adjusting channel dimensions.
$\text{Split}(\cdot)$ reshapes or partitions the input for grouped spectral processing.
$\mathcal{F}(\cdot)$ and $\mathcal{F}^{-1}(\cdot)$ denote the 2D Fast Fourier Transform and its inverse.
$\text{Filter}(\cdot)$ applies learnable frequency-domain filters to both real and imaginary components.
$\text{Polymerize}(\cdot)$ fuses spectral branches into unified spatial features.
$\text{DWC}(\cdot)$ is a depthwise separable convolution for spatial refinement.
$\text{Gelu}(\cdot)$ applies nonlinearity, and $\odot$ denotes element-wise gating.
$f_{\text{out}}$ is the final output via residual addition.

\begin{figure}[htbp]
\centerline{\includegraphics[scale=0.31]{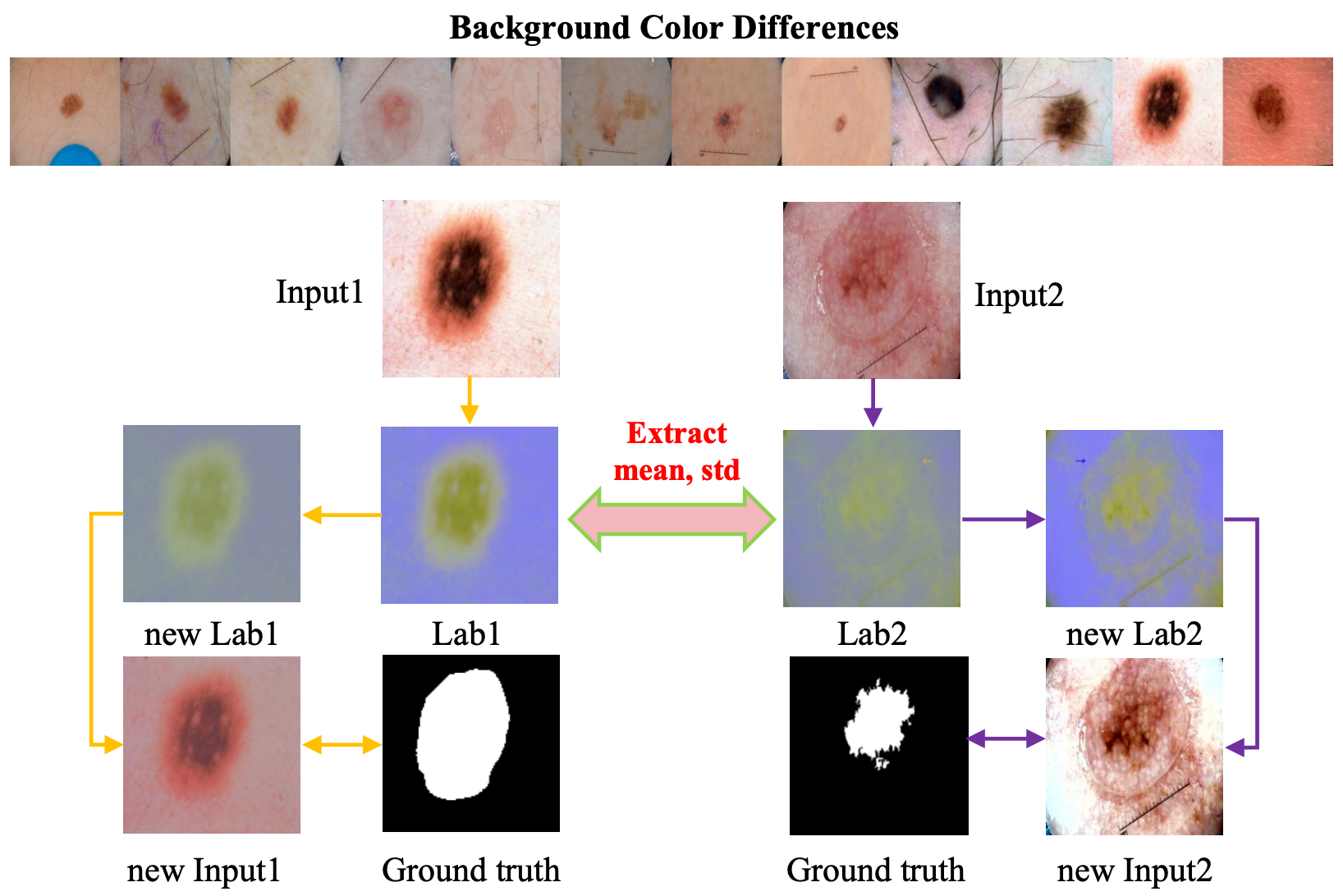}} 
\caption{Illustration of one training-time color transfer operation, where appearance is altered while preserving anatomical structure.}
\label{fig5}
\end{figure}

\subsection{Color Transfer for Robust Segmentation}

Color inconsistency is a common issue in medical images acquired under varying clinical conditions, particularly in modalities such as endoscopy, dermoscopy, or fundus photography. Differences in illumination, hardware settings, or patient factors can lead to distribution shifts that affect segmentation model robustness. To mitigate this, we apply a simple color transfer technique during training. For each input image, another image is randomly selected from the dataset, and its global color statistics are transferred in the LAB color space. By matching the per-channel statistics with a reference image in LAB space, we transfer its global color characteristics without altering the spatial structure. LAB space is used instead of RGB to decouple luminance and chromaticity, ensuring structural integrity and natural effect of the original image. Unlike heuristic augmentations such as ColorJitter, which introduce arbitrary distortions and can generate unnatural appearances, our approach (1) requires no handcrafted parameter tuning, (2) transfers color statistics from real in-domain images, and (3) generalizes well across different color-based medical imaging modalities.

The augmentation is used only during training. Original images are preserved at inference. As shown in Fig.~\ref{fig5}, this approach enables the same anatomy to appear under different color distributions without altering spatial layout. Algorithm~\ref{alg:color_transfer} summarizes the structure-preserving color transfer used in WaDiS. For each input image $I_s$, a reference image $I_r$ is randomly selected from the same dataset. Both images are converted to LAB color space, where luminance and chromaticity channels are independent, allowing controlled color manipulation. The mean and standard deviation of each channel in $I_r$ are used to normalize and re-scale $I_s$, thereby transferring the global color tone of the reference while preserving structural content. The resulting LAB image is converted back to RGB to obtain $\tilde{I}_s$, which retains anatomical integrity with natural color variation.

\begin{algorithm}[htbp]
\caption{Structure-Preserving Color Transfer}
\label{alg:color_transfer}
\begin{algorithmic}[1]
\Require Input image $I_s$, dataset $\mathcal{D}$
\Ensure Color-transferred image $\tilde{I}_s$
\State Randomly select a reference image $I_r \in \mathcal{D}, I_r \neq I_s$
\State Convert $I_s$ and $I_r$ from RGB to LAB color space
\State Compute channel-wise mean $\mu_s, \mu_r$ and standard deviation $\sigma_s, \sigma_r$
\State Apply color transfer: $I_s' = (I_s - \mu_s) / \sigma_s \cdot \sigma_r + \mu_r$
\State Clip values of $I_s'$ to $[0, 255]$
\State Convert $I_s'$ back to RGB to obtain $\tilde{I}_s$
\end{algorithmic}
\end{algorithm}

\subsection{Loss Function}

WaDiS is trained with a joint objective combining noise prediction and latent reconstruction within the diffusion framework. During training, the ground-truth mask $Y$ is encoded into its latent representation $x_0 = \psi(Y)$. Gaussian noise is progressively added over $T$ timesteps to obtain the noised latent $x_T$:
\begin{equation}
x_T = \sqrt{\bar{\alpha}_t}\,x_0 + \sqrt{1 - \bar{\alpha}_t}\,n,
\end{equation}
where $n \sim \mathcal{N}(0, I)$ and $\bar{\alpha}_t$ defines the noise schedule.

Conditioned on the wavelet-based embedding $x_c$, the denoising U-Net predicts the injected noise $\tilde{n} = f(x_T; x_c)$, with the corresponding loss:
\begin{equation}
\mathcal{L}_{noise} = \| \tilde{n} - n \|_1,
\end{equation}
where $\| \cdot \|_1$ denotes the L$_1$ norm, corresponding to the mean absolute error.

The latent reconstruction loss is defined as:
\begin{equation}
\tilde{x}_0 = \frac{1}{\sqrt{\bar{\alpha}_t}}\left(x_T - \sqrt{1 - \bar{\alpha}_t}\,\tilde{n}\right), 
\quad
\mathcal{L}_{latent} = \| \tilde{x}_0 - x_0 \|_1.
\end{equation}

The total objective is given by:
\begin{equation}
\mathcal{L}_{total} = \mathcal{L}_{noise} + \lambda \mathcal{L}_{latent},
\end{equation}
where $\lambda$ balances the two terms (set to 1 in practice). 
This joint loss encourages accurate noise estimation and structure-consistent latent reconstruction, improving segmentation precision and overall stability.

\section{Experiments}

\subsection{Datasets}
The effectiveness of our proposed method is validated across five diverse medical image segmentation benchmarks, each representing different imaging modalities. The selected datasets encompass colonoscopy, CT, dermoscopy, fundus, and ultrasound modalities, allowing a thorough assessment of segmentation performance across varied anatomical regions and heterogeneous imaging characteristics.

\textbf{CVC-ClinicDB}~\cite{bernal2015wm} comprises 612 colonoscopy images extracted from 31 video sequences, each accompanied by expert-annotated ground truth masks delineating polyp regions. All images have a native resolution of $384\times288$ pixels. This dataset serves as a standard benchmark for evaluating polyp segmentation performance under diverse endoscopic imaging conditions. The data are divided into training, validation, and test sets following an 8:1:1 ratio.

\textbf{Synapse}~\cite{ruan2024vm} is a publicly available abdominal CT dataset consisting of 30 contrast-enhanced cases with pixel-level annotations for eight abdominal organs, including the aorta, gallbladder, left kidney, right kidney, liver, pancreas, spleen, and stomach. The training set contains 2,211 slices, the validation set includes 1,568 slices, and the test set comprises 12 volumetric cases. Following our experimental setting, all annotated organs are treated collectively as a single foreground class to formulate a binary segmentation task.

\textbf{ISIC2018}~\cite{codella2019skin} is a publicly available dermoscopic image dataset from the International Skin Imaging Collaboration (ISIC) 2018 Challenge. It contains 2,694 dermoscopic images with expert-annotated segmentation masks of skin lesions. Following prior studies~\cite{ruan2022malunet}, the dataset is divided into 1,886 images for training and 808 for testing, corresponding approximately to a 7:3 split ratio.

\textbf{TN3K}~\cite{gong2021multi} dataset comprises 3,493 high-resolution ultrasound images collected from 2,421 patients, each containing at least one annotated thyroid nodule region. The dataset focuses on delineating thyroid nodules while excluding irrelevant color regions to ensure consistent grayscale appearance. Following a standardized clinical split protocol, 2,879 images are designated for training and 614 for testing.

\textbf{REFUGE2}~\cite{fang2022refuge2} dataset is a publicly available benchmark designed for glaucoma analysis, focusing on optic disc and cup segmentation in retinal fundus images. It contains 1,200 fundus images annotated independently by seven experienced ophthalmologists, each with an average of eight years of clinical experience. All annotations were subsequently reviewed and refined by a senior specialist with over ten years of expertise to ensure label consistency and clinical reliability. Following the setting of previous works~\cite{lin2024stable}, we combined the original training and validation subsets into a single training set of 800 images, with the remaining 400 images reserved for testing.


\subsection{Evaluation Metrics}

To comprehensively assess segmentation performance of our proposed method, we adopt three widely used quantitative metrics: the Dice Similarity Coefficient (DSC), Intersection over Union (IoU), and the 95th percentile Hausdorff Distance (HD95).

The Dice Similarity Coefficient (DSC) and Intersection-over-Union (IoU) are used to evaluate the region-level overlap between the predicted mask $\hat{Y}$ and the ground-truth mask $Y$. They are formally defined as:
\begin{equation}
\text{DSC} = \frac{2|\hat{Y} \cap Y|}{|\hat{Y}| + |Y|}, \quad
\text{IoU} = \frac{|\hat{Y} \cap Y|}{|\hat{Y} \cup Y|}.
\end{equation}
Both metrics measure pixel-wise agreement between prediction and ground truth, where higher values indicate better segmentation overlap and accuracy. To further evaluate boundary quality, we employ the 95th percentile Hausdorff Distance (HD95), which measures the spatial discrepancy between the predicted and reference contours (ground-truth mask). It is defined as:
\begin{equation}
\begin{split}
\text{HD}_{95}(A, B) = \operatorname{percentile}_{95} \Big(
& \{ \min_{b \in B} \| a - b \|_2 \}_{a \in A} \\
& \cup \{ \min_{a \in A} \| b - a \|_2 \}_{b \in B} 
\Big),
\end{split}
\end{equation}
where $A$ and $B$ denote the boundary point sets of the predicted mask $\hat{Y}$ and the ground-truth mask $Y$, respectively. A smaller HD95 value indicates more accurate boundary localization and reduced segmentation error near object contours.

In summary, DSC and IoU evaluate region-level consistency, while HD95 quantifies boundary accuracy, providing a more holistic assessment of segmentation performance in terms of both region overlap and contour precision.

\subsection{Implementation Details}

All experiments for WaDiS are conducted on a single NVIDIA RTX 3090 GPU with 24GB of memory. Training is conducted for 100{,}000 steps using AdamW with an initial learning rate of $1\times10^{-5}$ and a batch size set to 4. Latent diffusion operates in a compressed space with downsampling ratio $r = \frac{H}{h} = \frac{W}{w} = 8$, based on a KL-regularized autoencoder and U-Net denoiser. Input images are resized to $256 \times 256$, and their latent representations have spatial dimensions $32 \times 32$ with 4 channels. The WaDiS backbone is initialized from Stable Diffusion to leverage strong generative priors. Custom modules proposed in this work—namely the wavelet-based conditional encoder and the Frequency-Aware Gating Blocks (FAGB) are initialized with Xavier~\cite{glorot2010understanding}. Following SDSeg, newly added conditioning fusion layers in the denoising U-Net are zero-initialized to stabilize early training. At each training step, the color distribution of a randomly selected image is transferred to the input image via LAB-based transformation, altering its appearance while preserving spatial structure and semantic integrity. This augmentation is disabled during inference.

\subsection{Quantitative Analysis and Visualization}
To evaluate the segmentation performance of the proposed method on medical images, it is compared with several well-established segmentation models to demonstrate its effectiveness. Specifically, the comparative methods include CNN-based models such as U-Net~\cite{unet}, U-Net++~\cite{zhou2018unet++}, ResUNet~\cite{yu2019robust}, and DeepLabV3~\cite{chen2019rethinking}; Transformer-based models such as Swin-UNet~\cite{Swin-unet}, TransUNet~\cite{transunet}, HiFormer~\cite{heidari2023hiformer}, and MAXFormer~\cite{liang2023maxformer}; and Diffusion-based methods such as Diff-Trans~\cite{chowdary2023diffusion}, MedSegDiff-V2~\cite{wu2024medsegdiff}, LEAF~\cite{huang2025leaf}, and SDSeg~\cite{lin2024stable}. The default configurations and hyperparameter settings of all compared methods are adopted following their original implementations for a fair comparison.

\begin{figure*}[htbp]
\centerline{\includegraphics[width=\linewidth]{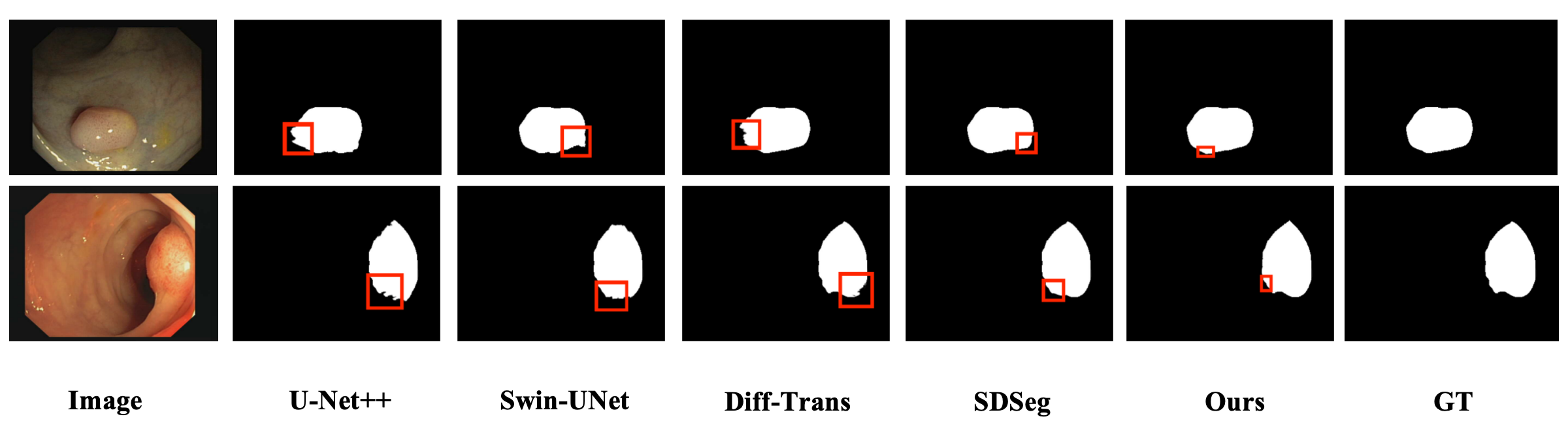}}
\caption{Qualitative comparison of segmentation results on CVC-ClinicDB dataset. Red boxes highlight discrepancies between Ground Truth (GT) and predictions from various baseline models.}
\label{fig:cvc_vis}
\end{figure*}

\subsubsection{Performance Comparison on CVC-ClinicDB Dataset}
We evaluate WaDiS on the CVC-ClinicDB dataset to verify its segmentation performance. As reported in Table~\ref{tab:cvc_results}, WaDiS achieves the highest scores across DSC and IoU metrics, while also attaining the lowest HD95, indicating more precise boundary localization. In particular, it outperforms SDSeg by a clear margin in both region-level and boundary-based metrics, validating the effectiveness of the wavelet-based conditional encoder and the frequency-aware gating blocks (FAGB). In addition, the use of color transfer augmentation during training improves robustness against color distribution shifts in clinical images. Qualitative results shown in Fig.~\ref{fig:cvc_vis} further demonstrate WaDiS’s ability to delineate anatomical boundaries while suppressing irrelevant background disturbance. Compared to other methods that often yield blurred or incomplete masks, WaDiS produces more accurate predictions.

\begin{table}[htbp]
\caption{Quantitative Comparison on CVC-ClinicDB Dataset. The highest-performing values are emphasized in \textbf{bold}.}
\label{tab:cvc_results}
\centering
\begin{tabular}{llccc} 
\hline
\textbf{Structure}&\textbf{Method} & \textbf{DSC$\uparrow$} & \textbf{IoU$\uparrow$} & \textbf{HD95$\downarrow$}\\
\hline
\multirow{4}{*}{CNN} 
&UNet~\cite{unet}   &85.12  &  81.32  & 10.83  \\
&UNet++~\cite{zhou2018unet++}  & 89.59  & 84.13  &  9.42  \\
&ResUNet~\cite{yu2019robust} &89.78 & 84.46 & 9.15\\
&DeepLabV3~\cite{chen2019rethinking} & 90.72 &86.19&8.73 \\
\hline
\multirow{4}{*}{Transformer} 
&Swin-Unet~\cite{Swin-unet}        & 92.42       & 88.69 & 7.42      \\
&TransUNet~\cite{transunet}   & 91.47 & 87.65 & 7.83\\
&Hiformer~\cite{heidari2023hiformer}  &93.29 & 88.95 &  6.98\\
&MAXFormer~\cite{liang2023maxformer}  &94.83 & 90.78 &6.31\\
\hline
\multirow{4}{*}{Diffusion} 
&Diff-Trans~\cite{chowdary2023diffusion}  & 95.40   & 92.00   &   5.95   \\
&MedSegDiff-V2~\cite{wu2024medsegdiff} & 94.85 & 91.79 & 6.02\\
&LEAF~\cite{huang2025leaf} & 94.95 & 92.12 & 6.12 \\
&SDSeg~\cite{lin2024stable}      & 95.80     & 92.60    &     5.77  \\
&\textbf{WaDiS (Ours)} & \textbf{96.33} & \textbf{92.88}  &  \textbf{5.31} \\
\hline
\end{tabular}
\end{table}

\subsubsection{Performance Comparison on Synapse Dataset}
\begin{figure*}[htbp]
\centerline{\includegraphics[width=\linewidth]{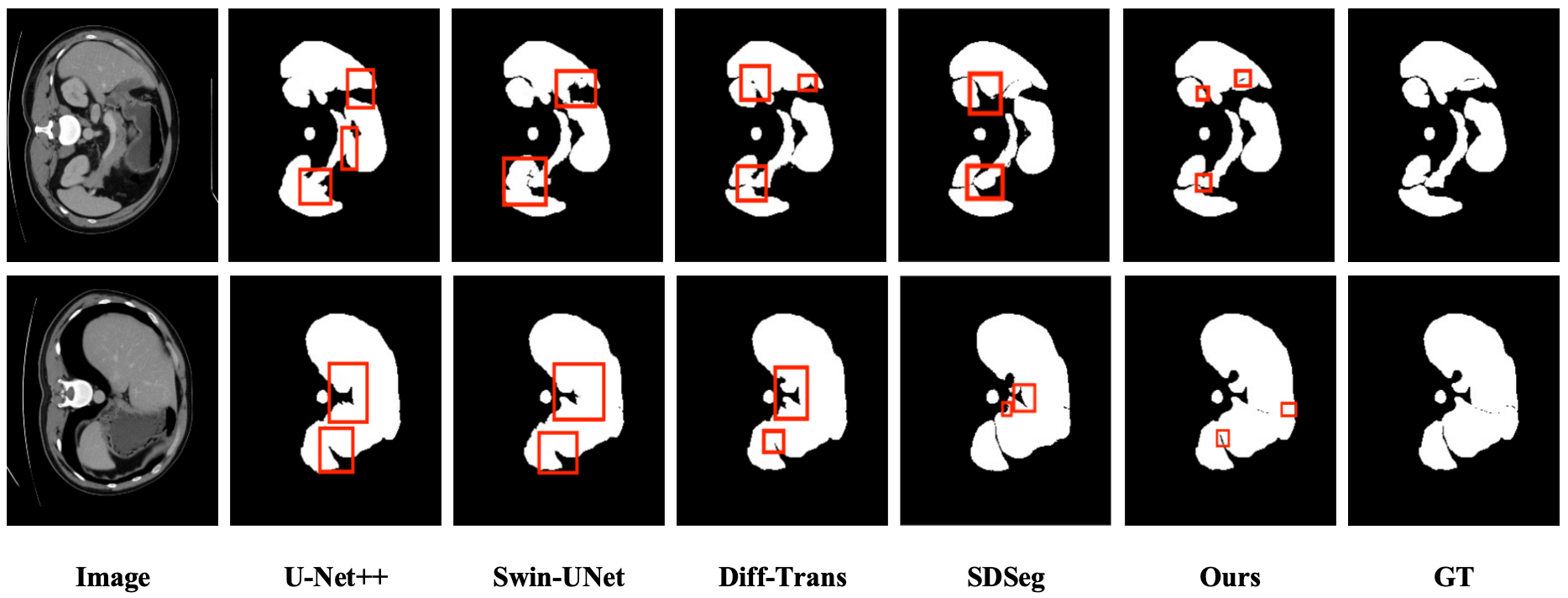}}
\caption{Qualitative comparison of segmentation results on Synapse dataset. Note that WaDiS performs binary segmentation, treating all organs as foreground and the rest as background. Red boxes highlight discrepancies between Ground Truth (GT) and predictions from various baseline models.}
\label{fig:syn_vis}
\end{figure*}

To evaluate the robustness of WaDiS under anatomical variation and low-contrast conditions, we test it on the Synapse multi-organ CT dataset, which includes diverse abdominal structures with varying shapes and boundaries. As shown in Table~\ref{tab:synapse_results}, WaDiS achieves the best performance in DSC and IoU, along with a competitive HD95, demonstrating strong region-level accuracy and reliable boundary precision. The improvements are particularly evident in structurally ambiguous or low-contrast regions, supported by the proposed multi-scale wavelet-based conditional encoding and frequency-aware gating blocks. Qualitative results in Fig.~\ref{fig:syn_vis} show that WaDiS produces sharper and more anatomically consistent boundaries, yielding complete and well-delineated organ contours even in challenging cases.

\begin{table}[htbp]
\caption{Quantitative Comparison on Synapse Dataset. The highest-performing values are emphasized in \textbf{bold}.}
\label{tab:synapse_results}
\centering
\begin{tabular}{llccc}
\hline
\textbf{Structure}&\textbf{Method} & \textbf{DSC$\uparrow$} & \textbf{IoU$\uparrow$}&\textbf{HD95$\downarrow$} \\
\hline
\multirow{4}{*}{CNN}
&UNet~\cite{unet}& 82.46 & 77.12 & 9.84\\
&UNet++~\cite{zhou2018unet++}   & 85.17   & 81.53  &  8.73    \\
&ResUNet~\cite{yu2019robust}& 85.68 & 82.01 & 8.42\\
&DeepLabV3~\cite{chen2019rethinking}& 86.10 & 82.35 & 7.95 \\
\hline
\multirow{4}{*}{Transformer}
&Swin-Unet~\cite{Swin-unet}    & 86.50       & 81.29  &  7.62  \\
&TransUNet~\cite{transunet}& 88.04 & 83.65 & 7.12  \\
&Hiformer~\cite{heidari2023hiformer}& 89.25 & 84.92 & 6.68 \\
&MAXFormer~\cite{liang2023maxformer}& 90.32 & 86.10 & 6.21 \\
\hline
\multirow{4}{*}{Diffusion} 
&Diff-Trans~\cite{chowdary2023diffusion}  & 91.11    & 85.03   & 5.87   \\
&MedSegDiff-V2~\cite{wu2024medsegdiff}& 92.34 & 86.02 & 5.62 \\
&LEAF~\cite{huang2025leaf}  &  92.67    &   85.97   &  5.58 \\
&SDSeg~\cite{lin2024stable}      & 92.80       & 86.26    &   \textbf{5.43} \\
&\textbf{WaDiS (Ours)} & \textbf{93.38} & \textbf{88.56}  & 5.51  \\
\hline
\end{tabular}
\end{table}

\begin{table}[htbp]
\caption{Quantitative Comparison on ISIC2018 Dataset. The highest-performing values are emphasized in \textbf{bold}.}
\label{tab:isic_results}
\centering
\begin{tabular}{llccc}
\hline
\textbf{Structure}&\textbf{Method} & \textbf{DSC$\uparrow$} & \textbf{IoU$\uparrow$}&\textbf{HD95$\downarrow$} \\
\hline
\multirow{4}{*}{CNN}
&UNet~\cite{unet}  & 84.23 & 77.12 & 8.43  \\ 
&UNet++~\cite{zhou2018unet++} & 87.46      & 79.89   &   7.84     \\
&ResUNet~\cite{yu2019robust} & 86.73 & 79.15 & 7.62  \\
&DeepLabV3~\cite{chen2019rethinking} & 88.02 & 80.38 & 7.12 \\
\hline
\multirow{4}{*}{Transformer}
&Swin-Unet~\cite{Swin-unet}   & 86.32 & 78.53 &  6.98    \\
&TransUNet~\cite{transunet}  & 88.95 & 81.62 & 6.42 \\
&Hiformer~\cite{heidari2023hiformer}  & 89.63 & 82.30 & 6.10  \\
&MAXFormer~\cite{liang2023maxformer}& 90.41 & 82.91 & 5.74\\
\hline
\multirow{4}{*}{Diffusion} 
&Diff-Trans~\cite{chowdary2023diffusion}   & 86.91   & 79.22   &  6.32   \\
&MedSegDiff-V2~\cite{wu2024medsegdiff} & 88.57 & 80.55 & 5.98   \\
&LEAF~\cite{huang2025leaf}    &  87.95   &  80.79   & 5.77   \\
&SDSeg~\cite{lin2024stable}   & 88.21     & 80.96 &  5.82    \\
&\textbf{WaDiS (Ours)} & \textbf{91.28} & \textbf{83.17} &  \textbf{5.35} \\
\hline
\end{tabular}
\end{table}

\subsubsection{Performance Comparison on ISIC2018 Dataset} 
\begin{figure*}[htbp]
\centerline{\includegraphics[width=\linewidth]{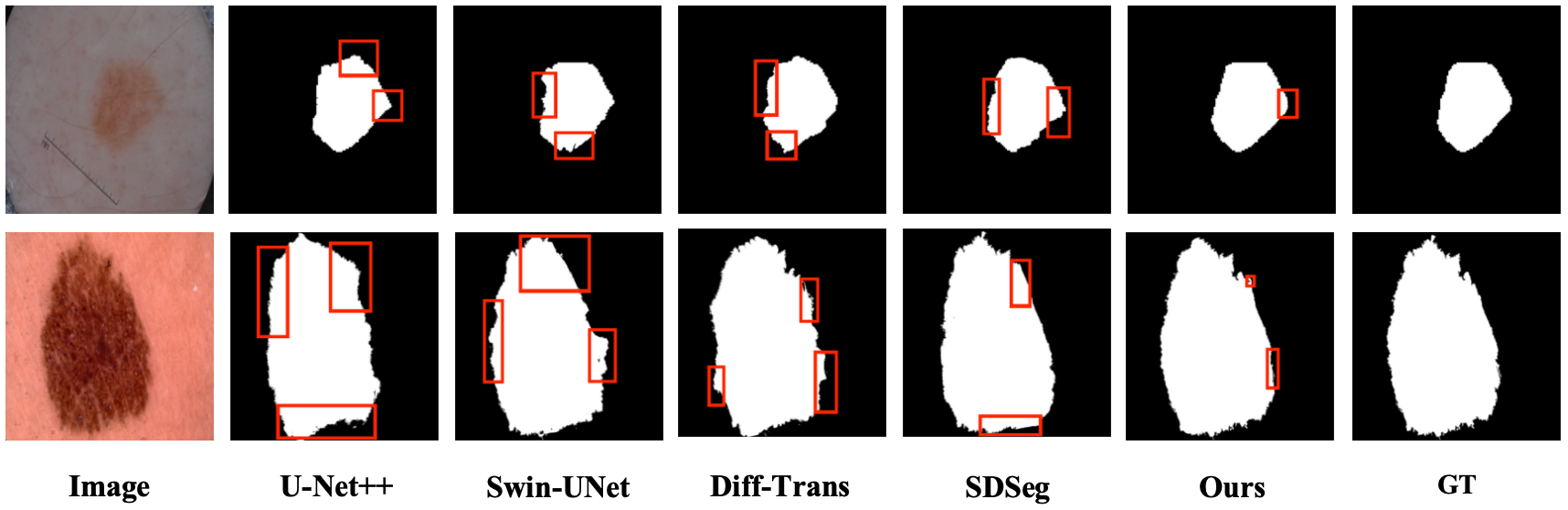}}
\caption{Qualitative comparison of segmentation results on ISIC2018 dataset. Red boxes highlight discrepancies between Ground Truth (GT) and predictions from various baseline models.}
\label{fig:isic_vis}
\end{figure*}

We further evaluate WaDiS on the ISIC2018 dataset, which features dermoscopic images with diverse lesion appearances and challenging low-contrast boundaries. As shown in Table~\ref{tab:isic_results}, WaDiS achieves the best overall performance across DSC, IoU, and HD95, demonstrating its superior capability in capturing irregular lesion contours and preserving fine boundary details. Representative results in Fig.~\ref{fig:isic_vis} demonstrate that WaDiS produces clearer and more accurate lesion boundaries than other methods, confirming its robustness and generalization to diverse skin lesion appearances.

\subsubsection{Performance Comparison on TN3K Dataset}
\begin{figure*}[htbp]
\centerline{\includegraphics[width=\linewidth]{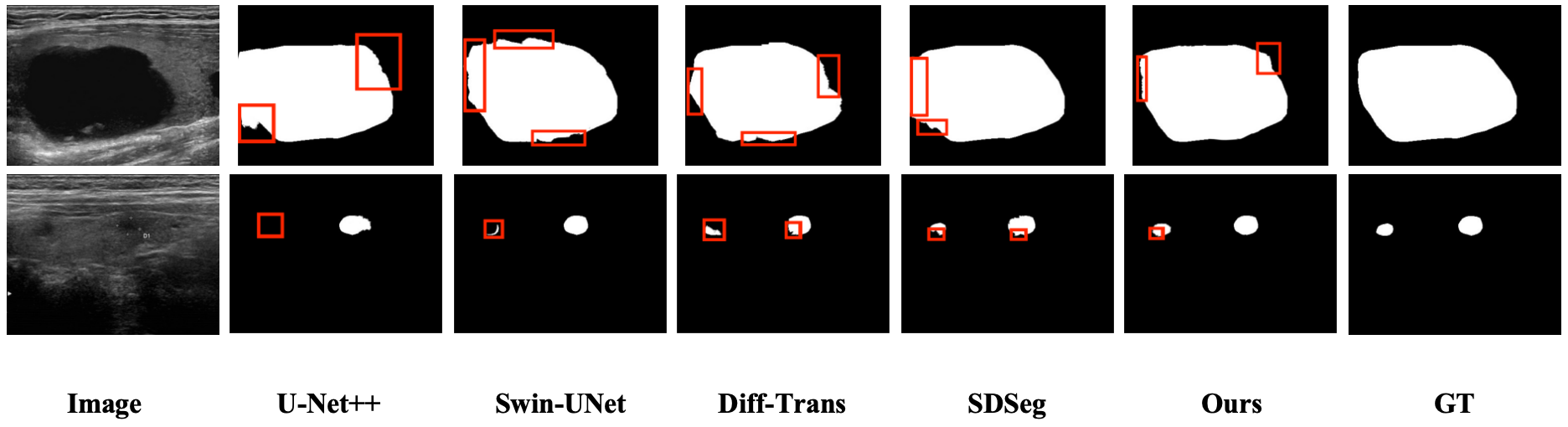}}
\caption{Qualitative comparison of segmentation results on TN3K dataset. Red boxes highlight discrepancies between Ground Truth (GT) and predictions from various baseline models.}
\label{fig:tn3k_vis}
\end{figure*}

To assess WaDiS on segmentation involving complex anatomical structures, we evaluate it on the TN3K ultrasound dataset, characterized by low-contrast regions with background noise and ambiguous boundary details. As shown in Table~\ref{tab:tn3k_results}, WaDiS achieves the best overall performance across DSC, IoU, and HD95 metrics, demonstrating improvements in both region-level accuracy and boundary precision. Performance gains are especially notable around object contours, where baseline models often exhibit blurred delineations or incomplete masks. Qualitative results in Fig.~\ref{fig:tn3k_vis} demonstrate that WaDiS generates masks with superior boundary sharpness and spatial coherence, confirming its robustness and generalization under challenging imaging conditions.

\begin{table}[htbp]
\caption{Quantitative Comparison on TN3K Dataset. The highest-performing values are emphasized in \textbf{bold}.}
\label{tab:tn3k_results}
\centering
\begin{tabular}{llccc}
\hline
\textbf{Structure}&\textbf{Method} & \textbf{DSC$\uparrow$} & \textbf{IoU$\uparrow$} &\textbf{HD95$\downarrow$}\\
\hline
\multirow{4}{*}{CNN}
&UNet~\cite{unet}   & 72.83 & 63.24 & 10.62 \\
&UNet++~\cite{zhou2018unet++}  & 75.56     & 65.89  & 9.84 \\
&ResUNet~\cite{yu2019robust} & 74.92 & 65.17 & 9.56\\
&DeepLabV3~\cite{chen2019rethinking} & 76.03 & 66.10 & 9.21 \\
\hline
\multirow{4}{*}{Transformer}
&Swin-Unet~\cite{Swin-unet} & 74.75 &  63.45 & 8.72\\
&TransUNet~\cite{transunet} & 76.89 & 67.21 & 8.35\\
&Hiformer~\cite{heidari2023hiformer} & 77.38 & 67.64 & 7.93 \\
&MAXFormer~\cite{liang2023maxformer} & 78.16 & 68.02 & 7.51\\
\hline
\multirow{4}{*}{Diffusion} 
&Diff-Trans~\cite{chowdary2023diffusion}  & 76.79  & 66.94 & 7.12 \\
&MedSegDiff-V2~\cite{wu2024medsegdiff}  & 78.05  & 68.05  & 6.85\\
&LEAF~\cite{huang2025leaf}        &  77.85  &  68.57  &  6.79  \\
&SDSeg~\cite{lin2024stable}    & 78.79   & 68.17   &  6.73    \\
&\textbf{WaDiS (Ours)} & \textbf{79.61} & \textbf{69.53}  & \textbf{6.41}  \\
\hline
\end{tabular}
\end{table}

\begin{table}[htbp]
\caption{Quantitative Comparison on REFUGE2 Dataset. The highest-performing values are emphasized in \textbf{bold}.}
\label{tab:refuge2_results}
\centering
\begin{tabular}{llccc}
\hline
\textbf{Structure}&\textbf{Method} & \textbf{DSC$\uparrow$} & \textbf{IoU$\uparrow$} &\textbf{HD95$\downarrow$}\\
\hline
\multirow{4}{*}{CNN}
&UNet~\cite{unet} & 86.12 & 78.65 & 5.73\\
&UNet++~\cite{zhou2018unet++}  & 87.84 & 80.10 & 5.42  \\
&ResUNet~\cite{yu2019robust}& 88.03 & 80.25 & 5.31 \\
&DeepLabV3~\cite{chen2019rethinking} & 88.55 & 80.74 & 5.12\\
\hline
\multirow{4}{*}{Transformer}
&Swin-Unet~\cite{Swin-unet} & 88.97 & 81.12 & 4.88\\
&TransUNet~\cite{transunet}& 89.35 & 81.46 & 4.67\\
&Hiformer~\cite{heidari2023hiformer}& 89.78 & 82.03 & 4.52\\
&MAXFormer~\cite{liang2023maxformer}& 90.12 & 82.40 & 4.38\\
\hline
\multirow{4}{*}{Diffusion} 
&Diff-Trans~\cite{chowdary2023diffusion} & 89.10 & 81.54 & 4.35 \\
&MedSegDiff-V2~\cite{wu2024medsegdiff}& 89.32 & 81.63 & 4.28\\
&LEAF~\cite{huang2025leaf}    &  89.34    &   80.99   &  4.31   \\
&SDSeg~\cite{lin2024stable}    &   89.40 &   81.80   &   4.20  \\
&\textbf{WaDiS (Ours)} & \textbf{91.20} & \textbf{84.02}  &  \textbf{3.82}\\
\hline
\end{tabular}
\end{table} 

\subsubsection{Performance Comparison on REFUGE2 Dataset}
\begin{figure*}[htbp]
\centerline{\includegraphics[width=\linewidth]{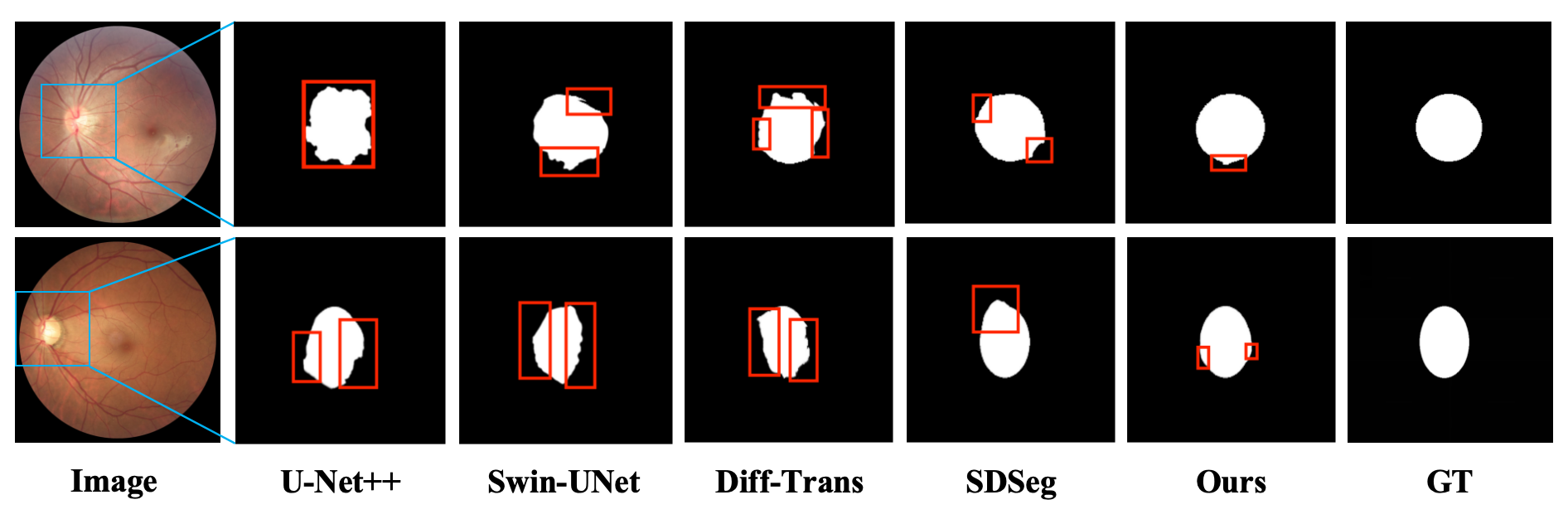}}
\caption{Qualitative comparison of segmentation results on REFUGE2 dataset. Red boxes highlight discrepancies between Ground Truth (GT) and predictions from various baseline models.}
\label{fig:refuge2_visual}
\end{figure*}

On the REFUGE2 dataset, WaDiS achieves the highest overall segmentation accuracy among all compared methods, demonstrating superior performance in DSC, IoU, and HD95, as shown in Table~\ref{tab:refuge2_results}. Compared with the best-performing baseline, SDSeg~\cite{lin2024stable}, WaDiS attains higher segmentation accuracy and boundary precision, benefiting from its wavelet-based conditional embedding and frequency-aware feature refinement. The Frequency-Aware Gating Block (FAGB) further enhances boundary-relevant representations through adaptive spectral filtering, leading to improved contour continuity in optic cup delineation. As illustrated in Fig.~\ref{fig:refuge2_visual}, WaDiS generates smoother and more anatomically faithful boundary contours, especially in areas where the transition between the optic cup and surrounding background is ambiguous. These results confirm that the proposed diffusion-based segmentation framework effectively balances semantic understanding and fine-grained boundary localization, achieving robust generalization across fundus images with complex appearance variations.

\begin{figure*}[!t]
    \centering
    \includegraphics[width=0.95\linewidth]{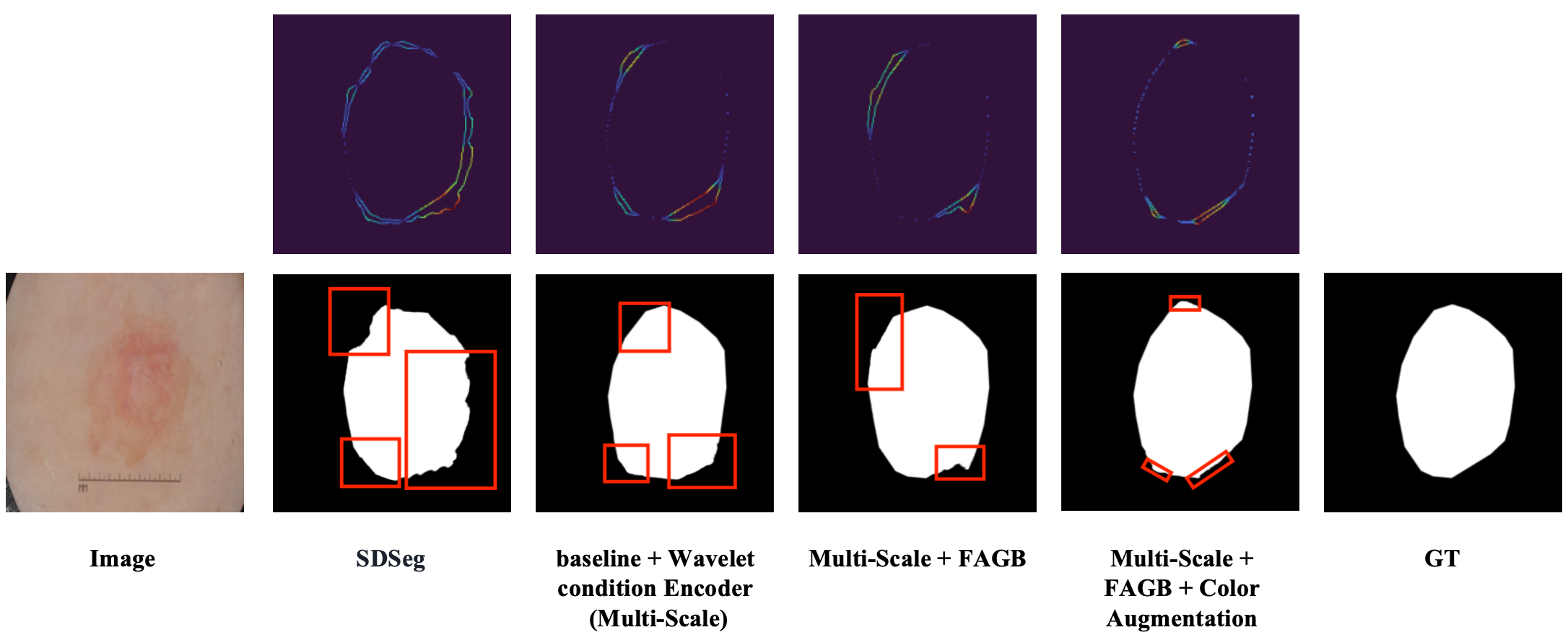}
    \caption{Qualitative ablation results on the ISIC2018. The top row presents the boundary error maps of different model variants, where color intensity indicates the degree of deviation from the ground truth boundary. The bottom row displays the input image, segmentation outputs from different model variants (SDSeg baseline, baseline with the wavelet-based conditional encoder, addition of the frequency-aware gating block (FAGB), and incorporation of LAB-based color augmentation), followed by the ground truth (GT). The visual comparison reveals that each component progressively reduces boundary errors and enhances contour completeness, leading to more anatomically consistent and stable segmentation outcomes.}
    \label{fig:vis_ablation}
\end{figure*}

\subsection{Ablation Study}

\begin{table}[htbp]
\centering
\caption{Ablation study on ISIC2018 dataset. The highest-performing values are emphasized in \textbf{bold}.}
\label{tab:ablation}
\begin{tabular}{lccc}
\hline
\textbf{Method Variant} & \textbf{DSC$\uparrow$} & \textbf{IoU$\uparrow$} &\textbf{HD95$\downarrow$}\\
\hline
SDSeg~\cite{lin2024stable} (Baseline)        & 88.21 & 80.96 & 5.82\\
+ Wavelet condition Encoder (Multi-Scale)    & 88.78  & 81.26 & 5.65\\
+ Multi-Scale + FAGB                        & 89.37 & 81.83 & 5.53 \\
+ Multi-Scale + FAGB + Color Augmentation   & \textbf{91.28} & \textbf{83.17} &\textbf{5.35}\\
\hline
\end{tabular}
\end{table}

To evaluate the effectiveness of individual components in WaDiS, we conduct a series of ablation experiments on the ISIC2018 skin lesion segmentation dataset. We begin with SDSeg~\cite{lin2024stable} as the baseline and incrementally introduce our proposed modules, including the wavelet-based condition encoder, the frequency-aware gating block (FAGB), and the LAB-based color augmentation. As summarized in Table~\ref{tab:ablation}, each component leads to a consistent improvement across all evaluation metrics, including DSC, IoU, and HD95. Replacing the CNN condition encoder with a multi-scale wavelet design leads to clear improvements, as hierarchical frequency decomposition offers richer structural cues for more effective semantic conditioning. By emphasizing structurally relevant features in the frequency domain, FAGB sharpens boundaries while mitigating noise and irrelevant activations. Finally, structure-preserving color augmentation boosts robustness to varying image color appearances during training. Qualitative results in Fig.~\ref{fig:vis_ablation} further illustrate the progressive improvements brought by each component. As more modules are introduced, the predicted segmentation masks present clearer lesion boundaries, improved structural completeness, and fewer segmentation inconsistencies, demonstrating that the proposed modules jointly enhance contour accuracy and overall segmentation performance.

\begin{table*}[!t]
	\setlength\tabcolsep{0.6pt}
	\renewcommand\arraystretch{0.25} 
	\small 
 \centering
\caption{The parameters, inference time, and computational complexity for each model were evaluated under identical conditions on a single NVIDIA GeForce RTX 3090 GPU using the ISIC2018 dataset.}
\begin{tabular}{ccccccccc} 
    \toprule 
     Models & U-Net \cite{unet} & DeepLabV3 \cite{chen2019rethinking} & TransUnet \cite{transunet} & Swin-Unet \cite{Swin-unet} &Diff-Trans~\cite{chowdary2023diffusion}& MedSegDiff-V2~\cite{wu2024medsegdiff} & SDSeg~\cite{lin2024stable}& Ours\\
    \midrule 
    Params (M) &  37.57 & 39.10 & 104.93 & 27.27 &43.32 & 46.00 & 41.74& 43.63\\
    Inference time (ms) &11.88&  17.35& 33.29 & 16.57 & 23.41 & 24.87 & 22.72  &23.58\\
    FLOPs (G) & 51.19 & 62.85 & 200.68 & 30.89 & 135.67 & 149.01 & 130.43 & 145.53\\
    \bottomrule 
\end{tabular}
\label{tab_para} 
\end{table*} 

\subsection{Discussion}
Results across multiple datasets indicate that the proposed WaDiS framework effectively balances segmentation accuracy and computational efficiency. As shown in Table~\ref{tab_para}, although WaDiS introduces slightly higher parameter count and FLOPs compared with SDSeg and Diff-Trans, it achieves consistently superior segmentation performance in terms of DSC, IoU, and HD95. The wavelet-based conditional encoder enhances multi-scale structural representation, allowing the model to retain fine details even under low contrast and uneven illumination. Meanwhile, the frequency-aware gating block (FAGB) selectively emphasizes high-frequency components relevant to anatomical boundaries, improving contour sharpness. These designs enable WaDiS to produce precise and anatomically consistent predictions across diverse imaging modalities.

However, the inclusion of frequency-domain operations inevitably increases computational overhead, as reflected in a moderate rise in inference time and FLOPs. Moreover, the current 2D architecture limits spatial continuity modeling, which may restrict its generalization to 3D volumetric modalities such as MRI or CT. Future work will focus on developing lightweight 3D wavelet representations and more efficient frequency-domain strategies to maintain segmentation precision while reducing computational complexity.

\section{Conclusion}
In this paper, we introduced WaDiS, a wavelet-powered latent diffusion framework for medical image segmentation. The model integrates a multi-scale wavelet-based encoder that preserves structural fidelity in conditional representation, a frequency-aware gating block designed for boundary preservation, and a structure-preserving color augmentation scheme applied during training. WaDiS achieves robust and accurate performance across five diverse medical imaging datasets: CVC-ClinicDB, Synapse, ISIC2018, TN3K, and REFUGE2. The framework consistently improves segmentation accuracy and preserves structural detail, especially under challenging imaging conditions characterized by low contrast or boundary ambiguity. Future work will focus on extending WaDiS to multi-class and 3D segmentation, and exploring more efficient frequency-domain modules to enhance scalability in clinical settings.


\bibliographystyle{IEEEtran}
\bibliography{ref.bib}


\end{document}